\documentclass{article}

\usepackage[preprint]{neurips_2025}

\usepackage[utf8]{inputenc} 
\usepackage[T1]{fontenc}    
\usepackage{url}            
\usepackage{booktabs}       
\usepackage{amsfonts}       
\usepackage{nicefrac}       
\usepackage{microtype}      
\usepackage{xcolor}         
\usepackage{amsmath}
\definecolor{pixnerd}{HTML}{f8ad9d}
\usepackage[pagebackref=true,breaklinks=true,colorlinks,bookmarks=false,urlcolor=pixnerd]{hyperref}
\usepackage[capitalize]{cleveref}
\crefname{section}{Sec.}{Secs.}
\Crefname{section}{Section}{Sections}
\Crefname{table}{Table}{Tables}
\crefname{table}{Tab.}{Tabs.}

\usepackage{times}
\usepackage{epsfig}
\usepackage{graphicx}
\usepackage{amssymb}

\usepackage{multirow}
\usepackage{booktabs}
\usepackage{float}
\usepackage{makecell}  
\usepackage{diagbox} 
\usepackage{multirow}
\usepackage{subfig}
\usepackage{bbding}
\usepackage{adjustbox}
\usepackage{wrapfig}
\usepackage{pifont}

\newcommand{\bs}{\boldsymbol}

\title{{\color{pixnerd}PixNerd}: {\color{pixnerd}Pix}el {\color{pixnerd}Ne}u{\color{pixnerd}r}al Field {\color{pixnerd}D}iffusion}

\author{%
  Shuai Wang\textsuperscript{1} \quad Ziteng Gao\textsuperscript{3} \quad Chenhui Zhu\textsuperscript{1} \quad Weilin Huang\textsuperscript{2} \quad Limin Wang\textsuperscript{1} \\
  \\
  $^1$Nanjing University \quad $^2$ByteDance Seed \quad $^3$National University of Singapore \\
  \\
  {\bf \url{https://github.com/MCG-NJU/PixNerd}} \\
  {\bf \url{https://huggingface.co/spaces/MCG-NJU/PixNerd}} \\ [0.2cm]
}

\begin{document}

\maketitle
\begin{center}
   \vspace{-2em}
    \centering
    \captionsetup{type=figure}
    \includegraphics[width=0.66\linewidth]{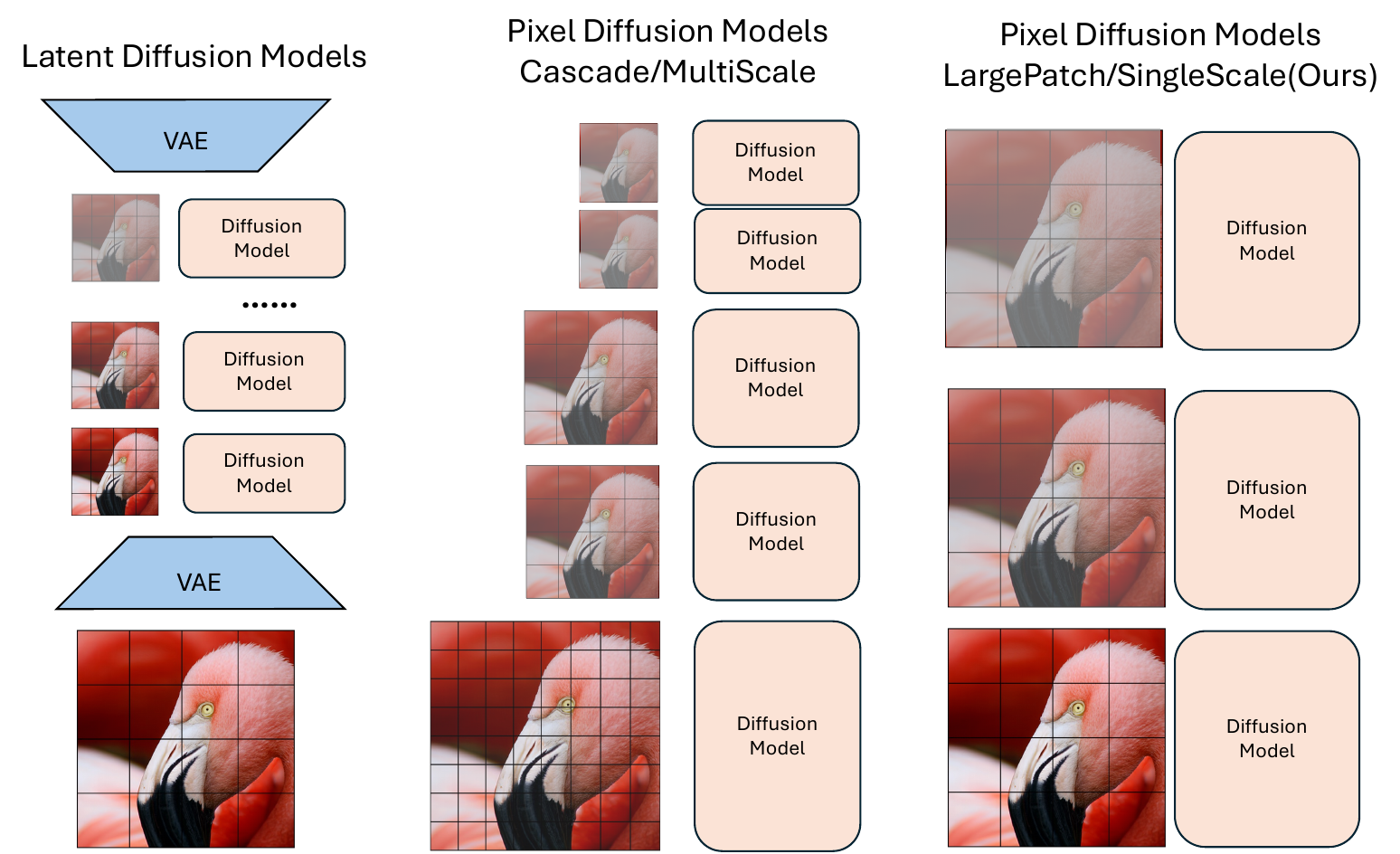}
    \includegraphics[width=0.30\linewidth]{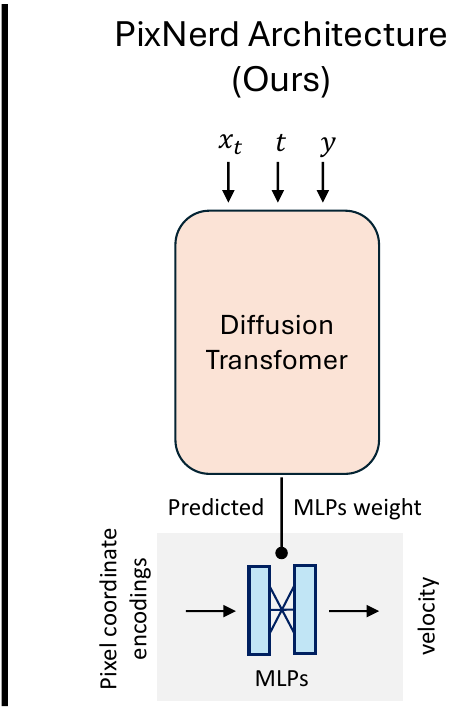}
    \caption{\textbf{Left: Comparison with other diffusion models.}  {\small Our LargePatch/SingleScale pixel space diffusion keeps consistent tokens as latent diffusion among diffusion steps.} \textbf{Right: PixNerd architecture.} {\small PixNerd follows the diffusion transformer design, replacing the final linear projection with a neural field to model the large patch details.}}
    \label{fig:main_arch}
    \vspace{-1em}
\end{center}
\begin{abstract}
The current success of diffusion transformers heavily depends on the compressed latent space shaped by the pre-trained variational autoencoder(VAE). However, this two-stage training paradigm inevitably introduces accumulated errors and decoding artifacts. To address the aforementioned problems, researchers return to pixel space at the cost of complicated cascade pipelines and increased token complexity. In contrast to their efforts, we propose to model the patch-wise decoding with neural field and present a single-scale, single-stage, efficient, end-to-end solution, coined as pixel neural field diffusion~(PixelNerd). Thanks to the efficient neural field representation in PixNerd, we directly achieved 2.15 FID on ImageNet $256\times256$ and 2.84 FID on ImageNet $512\times512$ without any complex cascade pipeline or VAE. We also extend our PixNerd framework to text-to-image applications. Our PixNerd-XXL/16 achieved a competitive 0.73 overall score on the GenEval benchmark and 80.9 overall score on the DPG benchmark. 
\vspace{-1em}
\end{abstract}
\section{Introduction}
\vspace{-0.5em}

\begin{figure}
    \centering
    \includegraphics[width=0.99\linewidth]{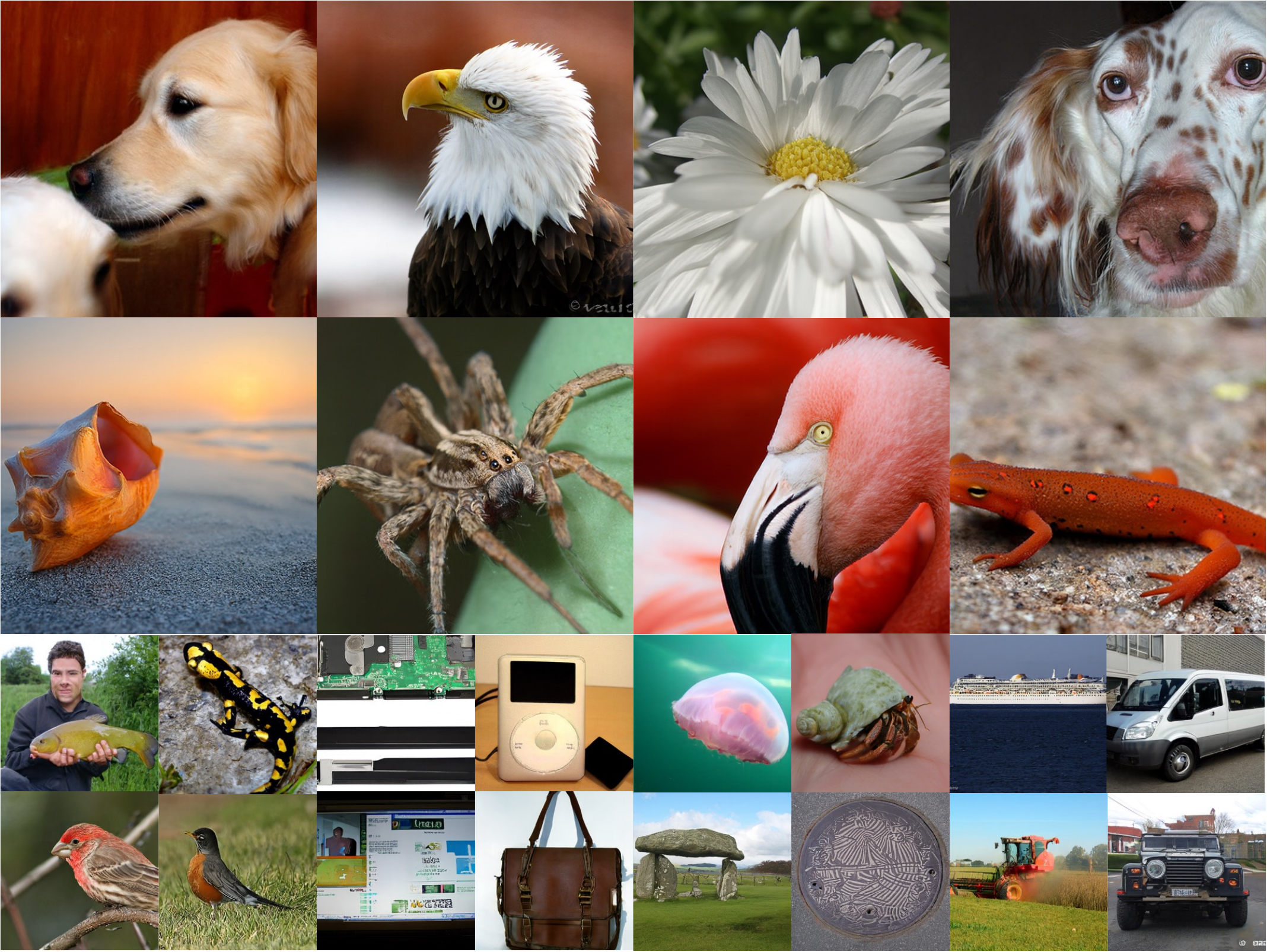}
    \caption{\textbf{Selected $256\times256$ and $512\times512$ resolution samples.} {\small Generated from PixNerd-XL/16 trained on ImageNet  $256\times256$ resolution and ImageNet $512\times512$ resolution with CFG = 3.5.}}
    \vspace{-2em}
    \label{fig:teaser}
\end{figure}
\begin{figure}
    \centering
    \includegraphics[width=0.99\linewidth]{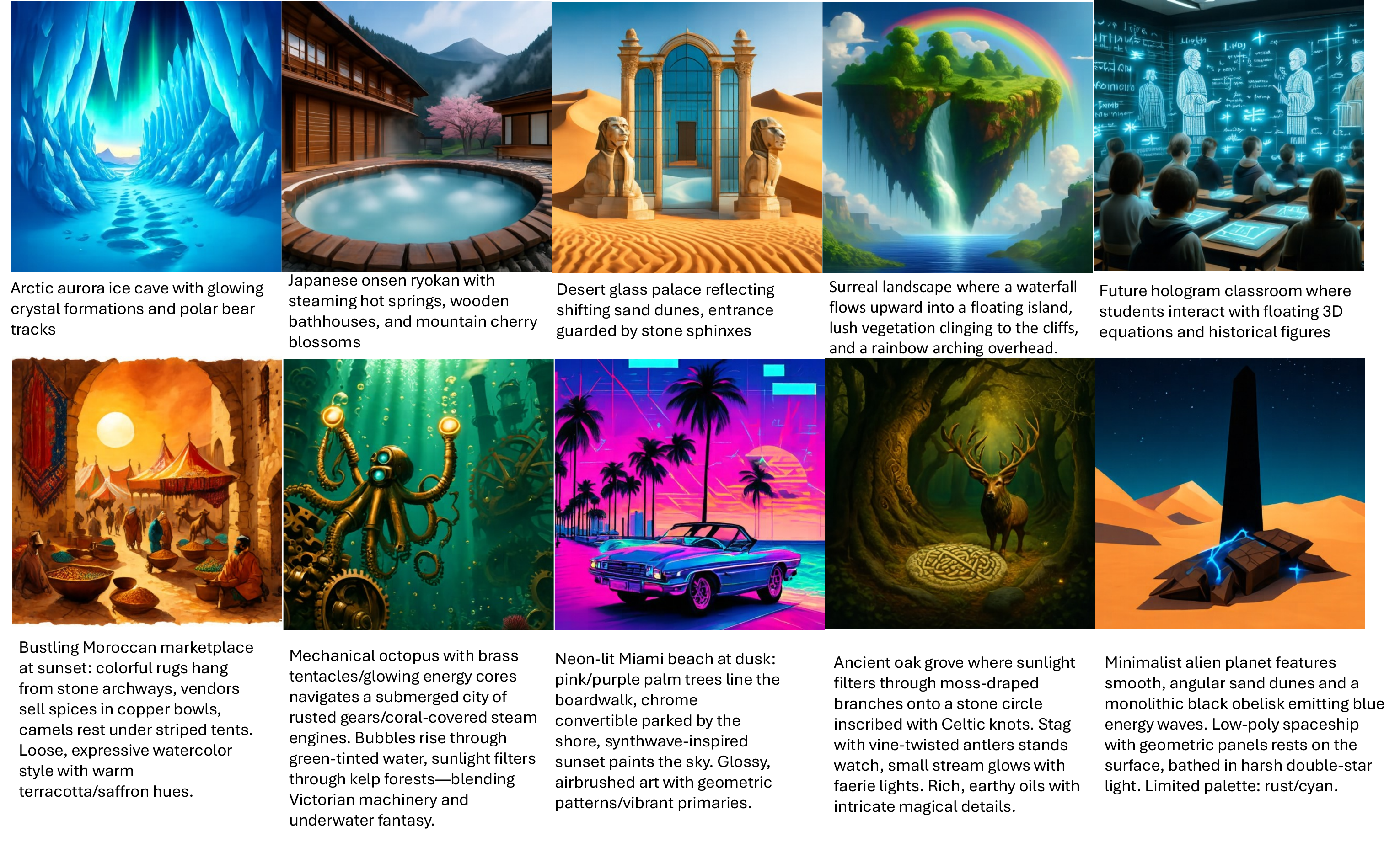}
    \caption{\textbf{The Text-to-Image $512\times512$ visualization with text descriptions of different lengths and styles.} {\small Given text descriptions of different lengths and styles, PixNerd can generate promising samples with a large patch size of 16. We used Adams-2nd solver with 25 steps and a CFG value of 4.0 for sampling.}}
    \label{fig:t2i_vis}
    \vspace{-2em}
\end{figure}

\begin{figure}
    \centering
    \includegraphics[width=0.99\linewidth]{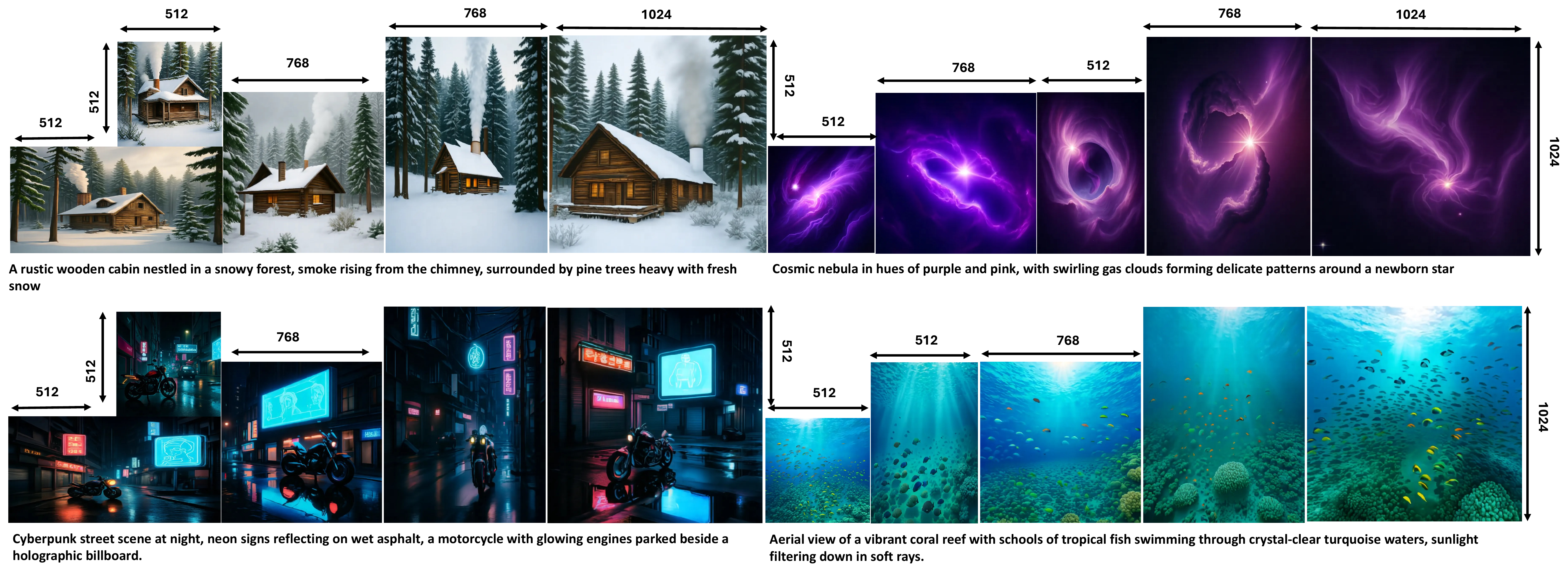}
    \caption{\textbf{Training-free arbitrary resolution generation.} {\small We keep the amount of tokens in PixNerd as constant as pretraining resolution, we \textbf{only interpolate the neural field coordinates} for different resolutions to yield multi-resolution images. }}
    \label{fig:t2i_vis_multires}
    \vspace{-2em}
\end{figure}

The current success of diffusion transformers largely depends on variational autoencoders (VAEs)~\cite{ldm, vavae, dcae}. VAEs significantly reduce the spatial dimension of raw pixels while providing a compact and nearly lossless latent space, substantially easing the learning difficulty for diffusion transformers. By operating in this compressed latent space, diffusion transformers can effectively learn either the score function or velocity of the diffusion process using small patch sizes. However, training high-quality VAEs typically requires adversarial training\cite{largegan, styleganxl, ldm, vavae, dcae} and additional supervision~\cite{lpips}, introducing complex optimization challenges. Moreover, this two-stage training paradigm leads to accumulated errors and inevitable decoding artifacts. To address these limitations, some researchers have explored joint training approaches, though these come with substantial computational costs~\cite{repa_e}.

An alternative approach involves implementing diffusion models directly in raw pixel space~\cite{adm, pixelflow, rdm}. In contrast to the success of latent diffusion transformers, progress in pixel-space diffusion transformers proves significantly more challenging. Without the dimensionality reduction provided by VAEs~\cite{vavae, ldm, dcae}, pixel diffusion transformers allocate substantially more image tokens~\cite{pixelflow}, requiring impractical computational resources when using the same patch size as latent diffusion transformers. To maintain a comparable number of image tokens, pixel diffusion transformers must employ much larger patch sizes during denoising training. However, due to the vastness of raw pixel space, larger patches make diffusion learning particularly difficult. Previous methods~\cite{pixelflow, rdm} have employed cascade solutions that divide the diffusion process across different scales to reduce computational costs. However, this cascade approach complicates both training and inference. In contrast to these methods, our work explores the performance upper bound using a large-patch diffusion transformer while maintaining the same token count and comparable computational requirements as latent diffusion models.

Inspired by the success of implicit neural fields in scene reconstruction~\cite{nerf, siren}, we propose modeling large-patch decoding using an implicit neural field. We replace the traditional diffusion transformer's linear projection with an implicit neural field, naming this novel pixel-space architecture PixelNerd~(\textbf{Pixel} \textbf{Ner}ual Field \textbf{D}iffusion). Specifically, we predict the weights of each patch's neural field MLPs using the diffusion transformer's last hidden states. For each pixel within a local patch, we first transform its local coordinates into coordinate encoding. This encoding, combined with the corresponding noisy pixel value, is then processed by the neural field MLPs to predict the diffusion velocity. Our approach significantly alleviates the challenge of learning fine details under large-patch configurations.

Compared to previous latent diffusion transformers and other pixel-space diffusion models, our end-to-end PixelNerd offers a simpler, more elegant, and efficient solution. For class-conditional image generation on ImageNet $256\times256$, PixelNerd-XL/16 achieves a competitive 2.15 FID and significantly better 4.55 sFID (indicating superior spatial structure) than PixelFlow~\cite{pixelflow, rdm}. On ImageNet $512\times512$, PixelNerd-XL/16 maintains comparable performance with 2.84 FID. For text-to-image generation, PixelNerd-XXL/16-$512\times512$ achieves 73.0 on the GenEval benchmark and 80.9 average score on DPG benchmark.

Our contributions are summarized as follows:
\begin{itemize}
    \item We propose a novel, elegant, efficient end-to-end pixel space diffusion transformer with neural field, deemed as PixNerd.
    \item For the class-to-image generation, on ImageNet $256\times256$, our PixNerd-XL/16 achieves a comparable 2.15 FID with similar computation demands as its latent counterpart. On ImageNet $512\times512$, our PixNerd-XL/16 achieves a comparable 2.84 FID with similar computation demands as its latent counterpart.
    \item For the text-to-image generation, our PixNerd-XXL/16 achieves a 0.73 overall score on the GenEval benchmark and 80.9 average score on DPG benchmark.
\end{itemize}
\section{Related Work}
\paragraph{Latent Diffusion Models} train diffusion models on a compact latent space shaped by a VAE~\cite{ldm}. Compared to raw pixel space, this latent space significantly reduces spatial dimensions, easing both learning difficulty and computational demands~\cite{ldm, vavae, dcae}. Thus VAE has become a core component in modern diffusion models~\cite{dit, sit, decoupled_dit, edm2, dod, flowdcn, dim, dmm}. However, VAE training typically involves adversarial training and perceptual supervision, complicating the overall pipeline. Insufficient VAE training can lead to decoding artifacts~\cite{sid}, limiting the broader applicability of diffusion generative models. Earlier latent diffusion models primarily focused on U-Net-like architectures. The pioneering work of DiT~\cite{dit} introduced transformers into diffusion models, replacing the traditionally dominant U-Net~\cite{uvit, adm}. Empirical results~\cite{dit} show that, given sufficient training iterations, diffusion transformers outperform conventional approaches without relying on long residual connections. SiT~\cite{sit} further validated the transformer architecture with linear flow diffusion.

\paragraph{Pixel Diffusion Models} have progressed much more slowly than their latent counterparts. Due to the vastness of pixel space, learning difficulty and computational demands are typically far greater than those of latent diffusion models~\cite{adm, vdm, simple_diffusion}. Current pixel-space diffusion models still rely on long residuals~\cite{rdm, adm}, limiting further scaling. Early attempts primarily split the diffusion process into chunks at different resolution scales to reduce computational burdens~\cite{pixelflow, rdm}. Pixelflow~\cite{pixelflow} uses the same denoising model across all scales, while Relay Diffusion~\cite{rdm} employs distinct models for each. However, this cascaded pipeline inevitably complicates both training and sampling. Additionally, training these models at isolated resolutions hinders end-to-end optimization and requires carefully designed strategies. FractalGen~\cite{fractal} constructs fractal generative models by recursively applying atomic generative modules, resulting in self-similar architectures that excel in pixel-by-pixel image generation. TarFlow~\cite{tarflow} introduces a Transformer-based normalizing flow architecture capable of directly modeling and generating pixels.

\section{Method}
\subsection{Preliminary}
\paragraph{Diffusion Models} gradually adds ${\bs x_0}$ with Gaussian noise $\epsilon$ to perturb the corresponding known data distribution $p(x_0)$ into a simple Gaussian distribution. The discrete perturbation function of each $t$ satisfies $\mathcal{N}({\bs x}_t|\alpha_t {\bs x}_0, \sigma_t^2 {\bs I})$, where $\alpha_t, \sigma_t > 0$. It can also be written as \cref{eq:ddpm}.
\begin{equation}
    {\bs x}_t = \alpha_t {\bs x}_\text{real} + \sigma_t {\bs \epsilon} . \label{eq:ddpm}
\end{equation}
Moreover, as shown in \cref{eq:forward_sde}, \cref{eq:ddpm} has a forward continuous-SDE description, where $f(t) = \frac{\mathrm{d}\log \alpha_t}{\mathrm{d}t}$ and $ g(t) = \frac{\mathrm{d} \sigma_t^2}{\mathrm{d} t} - (\frac{\mathrm{d}\log \alpha_t}{\mathrm{d} t}\sigma_t^2)$. \cite{reverse_sde} establishes a pivotal theorem that the forward
SDE has an equivalent reverse-time diffusion process as in \cref{eq:reverse_sde}, so the generating process is equivalent to solving the diffusion SDE. Typically, diffusion models employ neural networks and distinct prediction parametrization to estimate the score function $\nabla \log_x p_{{\bs x}_t}({\bs x}_t)$ along the sampling trajectory~\cite{vp, edm, ddpm}.
\begin{align}
     {d}{\bs x}_t &= f(t){\bs x}_t \mathrm{d}t + g(t) \mathrm{d}{\bs w} . \label{eq:forward_sde} \\
     {d}{\bs x}_t &= [f(t){\bs x}_t - g(t)^2\nabla_{\bs x} \log p({\bs x}_t)] dt + g(t) {d}{\bs w} .  \label{eq:reverse_sde}
\end{align}

VP \cite{vp} also shows that there exists a corresponding deterministic process \cref{eq:reverse_ode} whose trajectories share the same marginal probability densities of \cref{eq:reverse_sde}.
\begin{equation}
     {d}{\bs x}_t = [f(t){\bs x}_t - \frac{1}{2}g(t)^2\nabla_{\bs x_t} \log p({\bs x}_t)] {d}t . \label{eq:reverse_ode}
\end{equation}
{Rectified Flow Model} simplifies diffusion model under the framework of \cref{eq:forward_sde} and \cref{eq:reverse_sde}. Different from \cite{ddpm} introduces non-linear transition scheduling, the rectified-flow model adopts linear function to transform data to standard Gaussian noise. Instead of estimating the score function $\nabla_{\bs x_t} \log p_t({\bs x}_t)$, rectified-flow models directly learn a neural network $v_\theta(x_t, t)$ to predict the velocity field ${\bs v}_t = {d} {\bs x}_t =  ({\bs x}_\text{real} - {\bs \epsilon})$.
\paragraph{Diffusion Transformer} was introduced into diffusion models~\cite{dit} to replace the traditionally dominant UNet architecture~\cite{uvit,adm}. Empirical evidence demonstrates that given sufficient training iterations, diffusion transformers outperform conventional approaches even without relying on long residual connections. SiT~\cite{sit} further validated the transformer architecture with linear flow diffusion. Given a noisy image latent ${\bs x}_t$ as \cref{eq:ddpm}, ${\bs y}$ is the condition, $t$ is the timestep, we first partition it into non-overlapping patches, converting it into a 1D sequence. These noisy patches are then processed through stacked self-attention and FFN blocks, with class label conditions incorporated via AdaLN modulation. Finally, a simple linear projection decodes the feature patches into either patch-wise score or velocity estimates:
\begin{align}
    \label{eq:block}
    \mathbf{X}_t &= \mathbf{X}_t + \text{AdaLN}({\bs y}, {\bs t}, \text{Attention}(\mathbf{X}_t)),\\
    \mathbf{X}_t &= \mathbf{X}_t + \text{AdaLN}({\bs y}, {\bs t}, \ \ \text{FFN}(\mathbf{X}_t)).
\end{align}
Recent architectural improvements such as SwiGLU~\cite{llama1,llama2}, RoPE~\cite{rope}, and RMSNorm~\cite{llama1, llama2} have been extensively validated in the research community~\cite{visionllama, vavae, fit, decoupled_dit, seedream2, seedream3, mogao}. 
\paragraph{Neural Field} is usually adopted to represent a scene through MLPs that map coordinates encodings to signals~\cite{nerf, siren, ddmi, coco_gan, infd}. It has been widely applied to objects~\cite{mipnerf, neural_volumes} and surface reconstruction\cite{neus, monosdf, neus2}. Specifically, recall that an MLP consists of a Linear, SiLU, and another Linear, we regard $\mathbf{W}^n_1$ as the weight for the first linear layer in MLP, while $\mathbf{W}^n_2$ is for the second. If we have a neural field with a single MLP $\{ \mathbf{W}_1, \mathbf{W}_2\}$ for naive 2D scene, given the query coordinate $(i, j)$, the coordinate will be transformed into cosine/sine encodings in \cref{eq:sine_encodings} or DCT-basis encodings in \cref{eq:dct_encodings}:
\begin{equation}
    \text{PE}(i, j) = \sin(2^0\pi i), \cos(2^0\pi i), ..., \sin(2^L\pi i), \cos(2^L\pi i), ....  \sin(2^L\pi j), \cos(2^L\pi j) \label{eq:sine_encodings}
\end{equation}
Then this encoding feature will be fed into neural field MLPs to extract features $\mathbf{V}^n(i,j)$ as \cref{eq:nerf}:
\begin{equation}
    \mathbf{V}^n(i,j) = \text{MLP}((\text{PE}(i, j))|\{ \mathbf{W}_1, \mathbf{W}_2\}) . \label{eq:nerf}
\end{equation}
Finally, $\mathbf{V}^n(i,j)$ can be used to regress the needed value, eg. RGB~\cite{nerf, siren}, density and SDF~\cite{monosdf}.

\subsection{Diffusion Transformer with Patch-wise Neural Field}
While VAEs~\cite{ldm, vavae, dcae} significantly reduce spatial dimensions in the latent space, rendering a single linear projection sufficient for velocity modeling in latent diffusion transformers, pixel diffusion transformers must handle substantially larger patch sizes to maintain computational parity with their latent counterparts. Under such conditions, a simple linear projection becomes inadequate for capturing fine details.

To address the limitations of linear projection, we propose modeling patch-wise velocity decoding using an implicit neural field. Formally, given the last hidden states $\mathbf{X}^n$ of the $n$-th patch in diffusion transformer, we predict neural field parameters $\{ \mathbf{W}^n_1 \in \mathbb{R}^{D_2 \times D_1}, \mathbf{W}^n_2 \in \mathbb{R}^{D_1 \times D_2}  \}$ from $\mathbf{X}^n$. 
\begin{equation}
    \mathbf{W}_1^n, \mathbf{W}_2^n = \text{Linear}(\text{SiLU}(\mathbf{X}^{n})) .
\end{equation}

To decode the pixel-wise velocity $\mathbf{v}^n(i,j)$ for the pixel coordinate $(i, j)$ in the $n$-th patch feature $\mathbf{X}^n$, where $i, j \in (0, K)$, we first encode the coordinates into encodings. These encodings $(\text{PE}(i, j)$, along with the noisy pixel value ${\bs x}^{n}(i, j))$, are then fed into the neural field MLP to predict the velocity. To enhance performance and stabilize training, we apply row-wise normalization to the neural field parameters. For brevity, we omit the timestep subscript here.
\begin{equation}
    \mathbf{V}^n(i,j) = \text{MLP}( \text{Concat}([\text{PE}(i, j), {\bs x}^{n}(i, j)]) ~~| ~\{ {\mathbf{W}^n_1 \over {||\mathbf{W}^n_1||}}, {\mathbf{W}^n_2 \over {||\mathbf{W}^n_2||}}\}~~) . 
\end{equation}
Finally, as shown in \cref{eq:v_decode} the pixel velocity $\mathbf{v}^n(i,j)$ is decoded from $\mathbf{V}^n(i,j)$ through a linear projection:
\begin{equation}
     \mathbf{v}^n(i,j)  = \text{Linear}(\mathbf{V}^n(i,j)) . \label{eq:v_decode}
\end{equation}

\section{Experiments}
We conduct ablation studies and baseline comparison experiments on ImageNet-$256\times256$. For class-to-image generation, we provide system-level comparisons on ImageNet-$256\times256$ and ImageNet-$512\times512$, and report FID~\cite{fid}, sFID~\cite{sfid}, IS~\cite{is}, Precision, and Recall~\cite{pr_recall} as the main metrics. For text-to-image generation, we report results collected on the GenEval~\cite{geneval} and DPG~\cite{dpg} benchmarks.
\vspace{-0.5em}
\paragraph{Training Details.} Inspired by recent advances in training schedulers~\cite{sd3}, architecture design~\cite{visionllama, vavae, fit, decoupled_dit}, and representation alignment~\cite{repa}, we incorporate SwiGLU, RMSNorm\cite{visionllama, llama1, llama2}, lognorm sampling\cite{sd3}, and representation alignment from DINOv2~\cite{dinov2, repa, decoupled_dit} to enhance our PixNerd model. Specifically, we add an additional representation alignment loss with a weight of $0.5$ to align the intermediate features from the 8th layer of our PixNerd model with the features extracted by DINOv2-Base.

\begin{table}
\centering
\small
\renewcommand\arraystretch{1.0}
\setlength{\tabcolsep}{1.5pt} 
\begin{tabular}{c | c c c | c c}
\toprule
 & \multicolumn{3}{c|}{\textbf{Inference}} & \multicolumn{2}{c}{\textbf{Training}} \\
Model & 1 image & 1 step & Mem (GB) & Speed (s/it) & Mem (GB) \\
\midrule
SiT-L/2(VAE-f8) & 0.51s & 0.0097s & 2.9 & 0.30 & 18.4 \\
\midrule
Baseline-L/16 & 0.48s & 0.0097s & 2.1 & 0.18 & 18 \\
PixNerd-L/16 & 0.51s & 0.010s & 2.1 & 0.19 & 22 \\
\midrule
ADM-G & 4.21s & 0.08s & 2.23 & / & / \\
PixelFlow-XL/4 & 10.1s & 0.084s$^\dagger$ & 4.0 & / & / \\
PixNerd-XL/16 & 0.65s & 0.012s & 3.1 & 0.27 & 33.9 \\
\bottomrule
\end{tabular}
\caption{\textbf{The resource consumption comparison.} {\small $^\dagger$ means the average time consumption for a single step across different stages. Our PixNerd consumes much less memory and latency~(Nearly $8\times$ fatser than other pixel diffusion models.} \label{tab:resource}}
\vspace{-2em}
\end{table}

\paragraph{Resource Consumption.} We employ \textbf{torch.compile} to optimize memory allocation and reduce redundant computations for both the baseline and PixNerd. As shown in \cref{tab:resource}, compared with latent counterparts, our PixNerd-L/16 achieves much higher training throughput without VAE latency and with similar inference memory. Compared to other pixel-space diffusion models, our PixNerd consumes significantly less memory and has lower latency (nearly $8\times$ faster than ADM-G~\cite{adm} and PixelFlow~\cite{pixelflow}). 

\subsection{Comparison with Baselines} 
\begin{figure}
    \centering
    \includegraphics[width=0.99\linewidth]{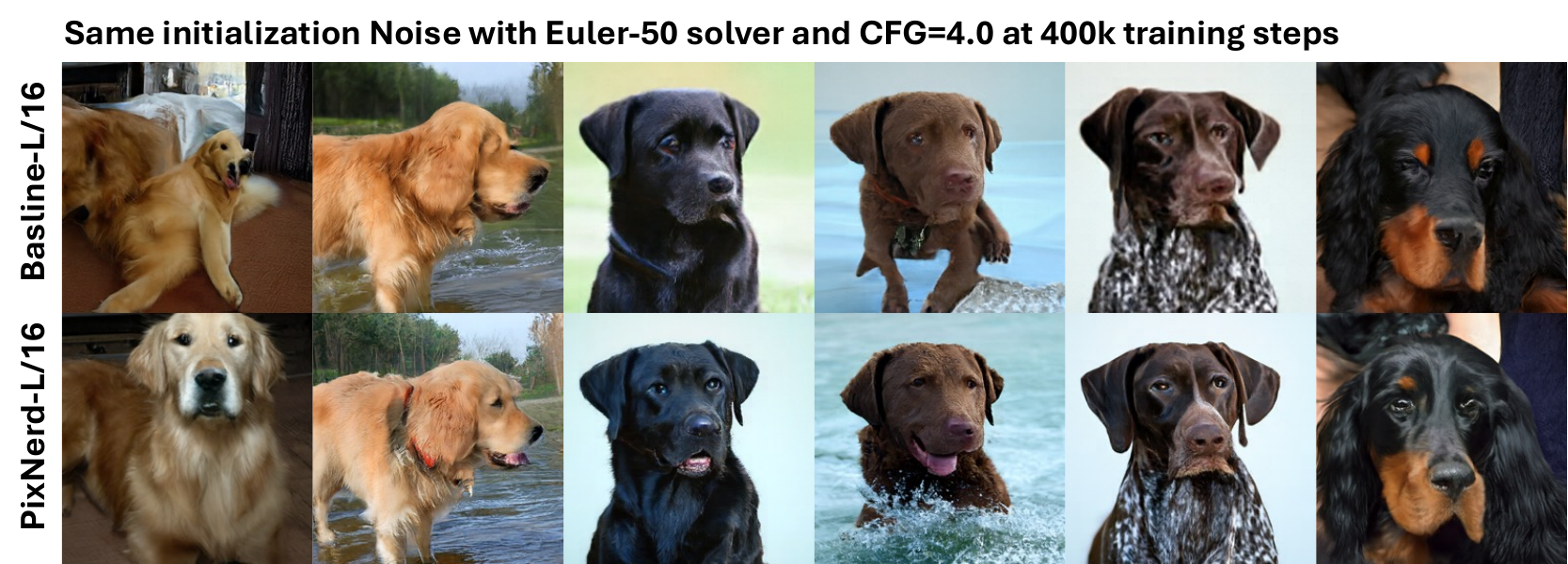}
    \caption{\textbf{The visualization Comparison with Baseline-L/16 under 400k training steps.}{\small With the help of neural field representation, our PixNerd-L/16 yields promising details and better structure.}}
    \label{fig:baseline_vis}
    \vspace{-2em}
\end{figure}

\begin{figure}
    \centering
   \subfloat[REPA loss~(DINOv2-B)\label{fig:baseline_repa}]{
        \includegraphics[width=0.48\linewidth]{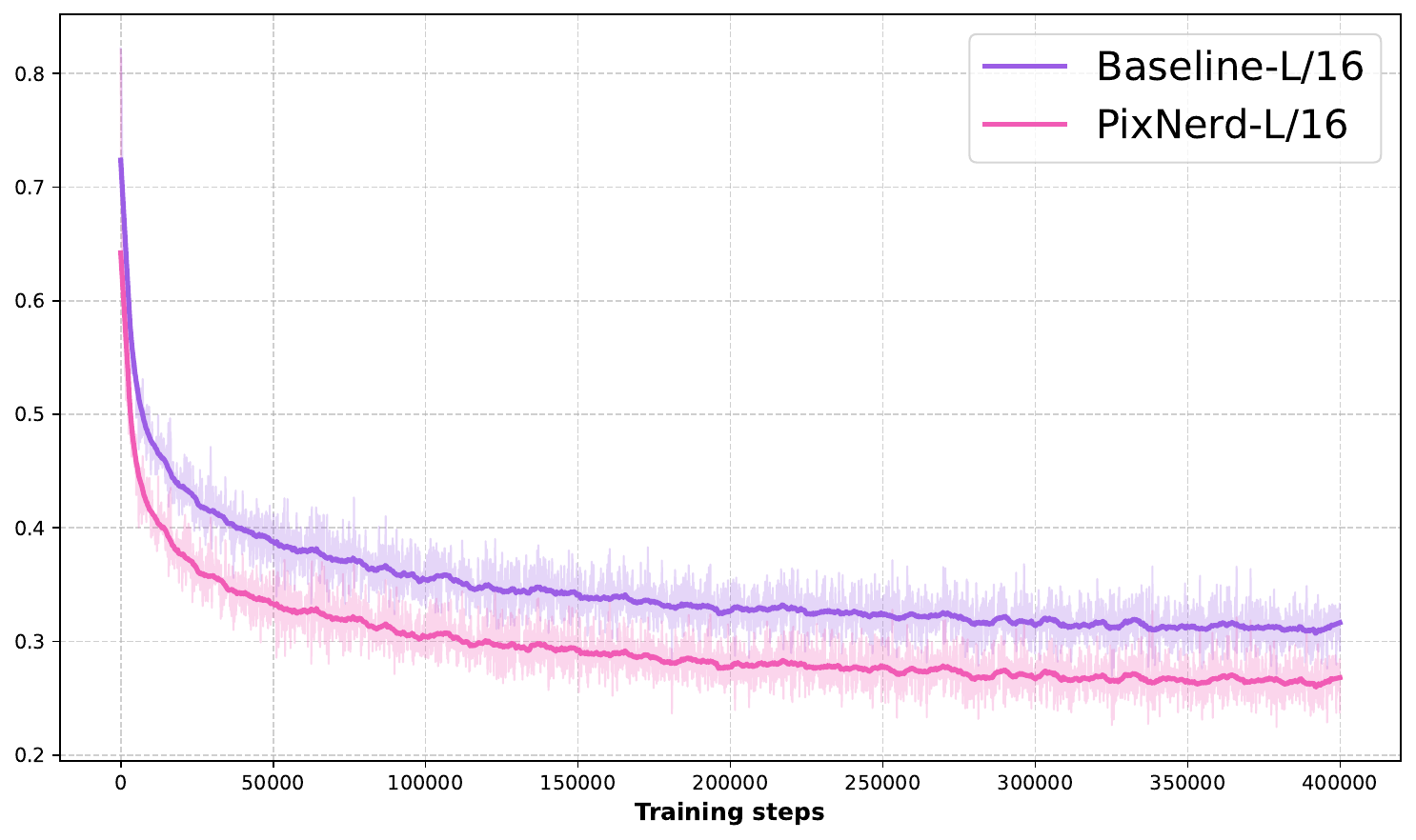}
    } 
    \subfloat[Flow Matching Loss\label{fig:baseline_flow}]{
        \includegraphics[width=0.48\linewidth]{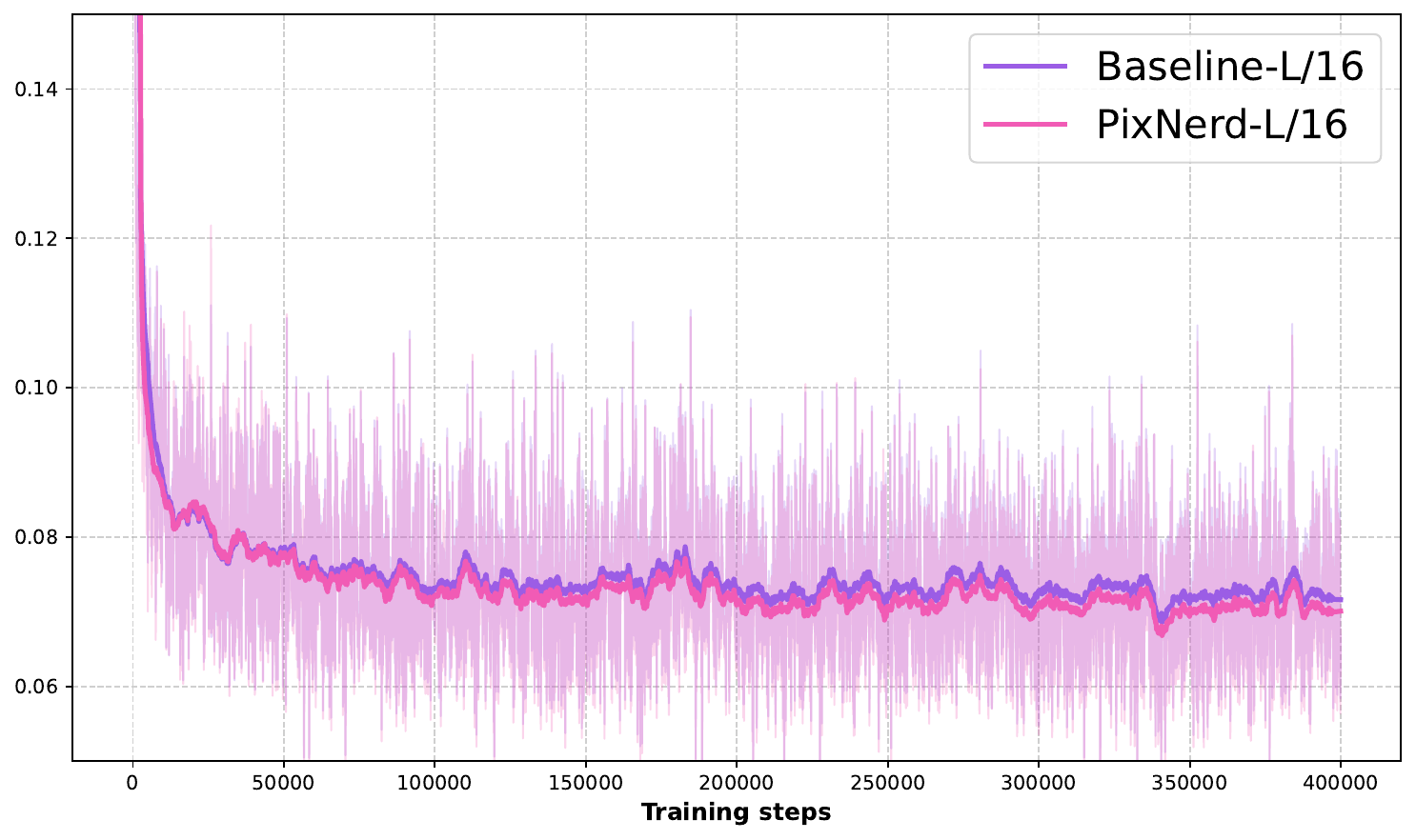}
    } 
    \caption{\textbf{Loss Comparison with Diffusion Transformer Baselines.} {\small Our PixNerd-L/16 achieves consistently lower REPA loss and flow matching loss than its diffusion transformer counterpart.}}
    \vspace{-2em}
\end{figure}

We conduct a baseline comparison on ImageNet $256\times256$ with a large-size model. Both Baseline-L/16 and PixNerd-L/16 are built upon SwiGLU~\cite{llama2, llama1}, RoPE2d~\cite{rope}, RMSNorm, and trained with lognorm sampling. All optimizer configurations are consistently aligned. As shown in \cref{fig:baseline_flow} and \cref{fig:baseline_repa}, our model achieves consistently lower loss expectation. We also provide visualization comparisons with Baseline-L/16 in \cref{fig:baseline_vis}: PixNerd-L/16, trained for the same number of steps, generates better details and structures.

\subsection{Neural Field Design}
We conduct ablation studies on PixNerd-L/16, which comprises 22 transformer layers with 1024 channels. The Neural Field design is configured to have a computational burden comparable to that of a two-layer transformer block, so PixNerd-L/16 has inference latency similar to its counterpart, Baseline-L/16~(24 transformer layers with 1024 channels).
\begin{figure}
    \centering
   \subfloat[Neural Field Normalization\label{fig:nerf_norm}]{
        \includegraphics[width=0.32\linewidth]{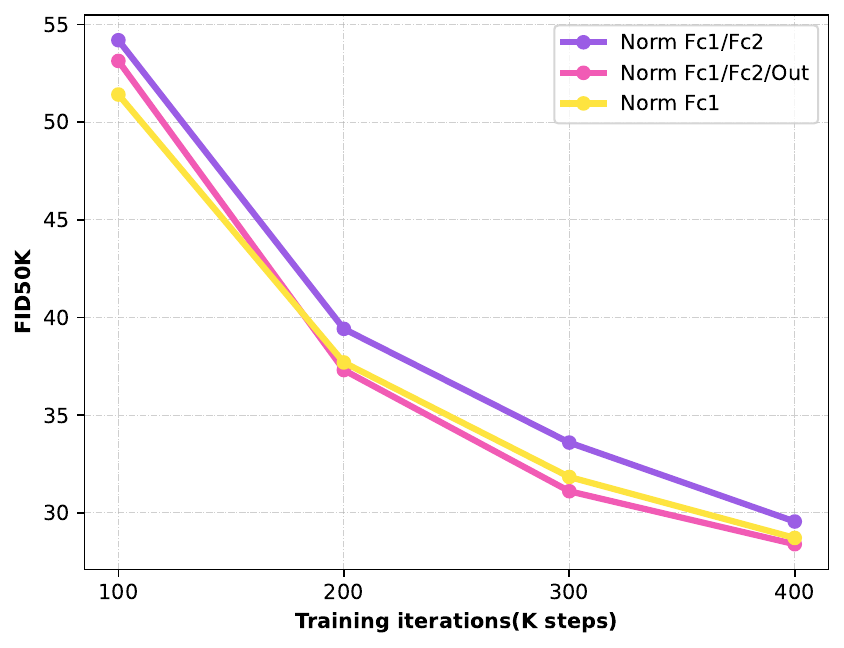}
    } 
    \subfloat[Neural Field Channels\label{fig:nerf_channels}]{
        \includegraphics[width=0.32\linewidth]{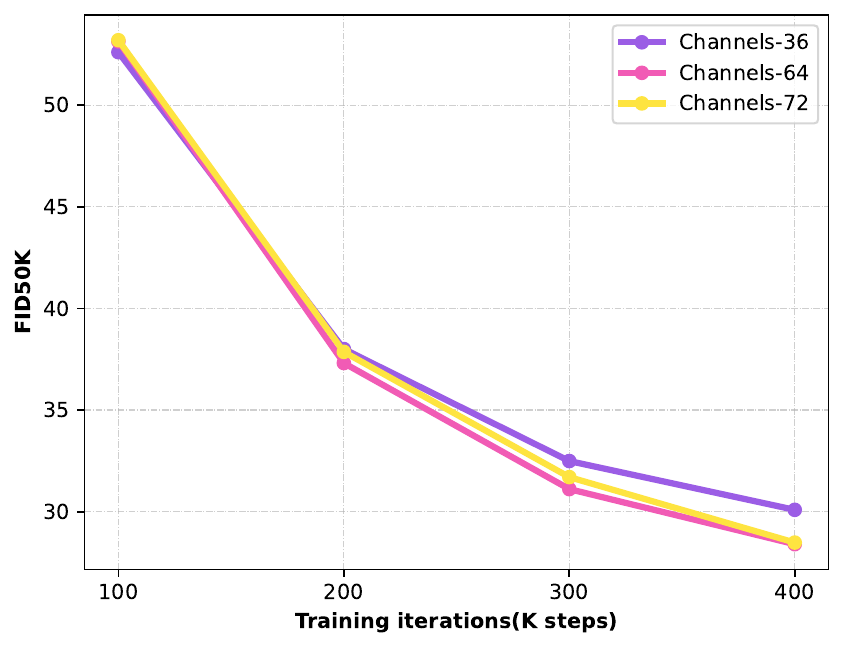}
    } 
    \subfloat[Neural Field MLPs layers\label{fig:nerf_depth}]{
        \includegraphics[width=0.32\linewidth]{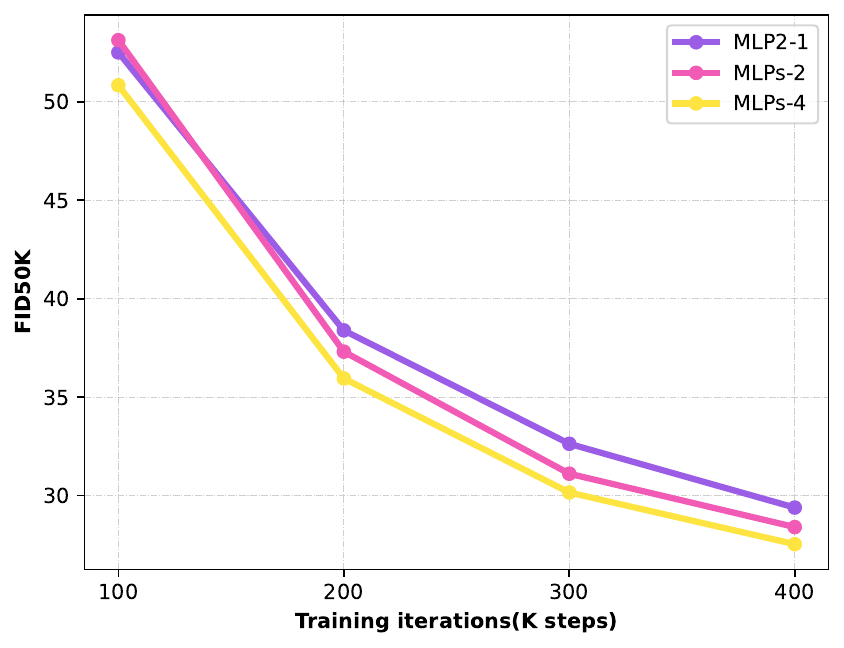}
    } 

    \subfloat[Coordinate-Encoding\label{fig:nerf_pe}]{
        \includegraphics[width=0.32\linewidth]{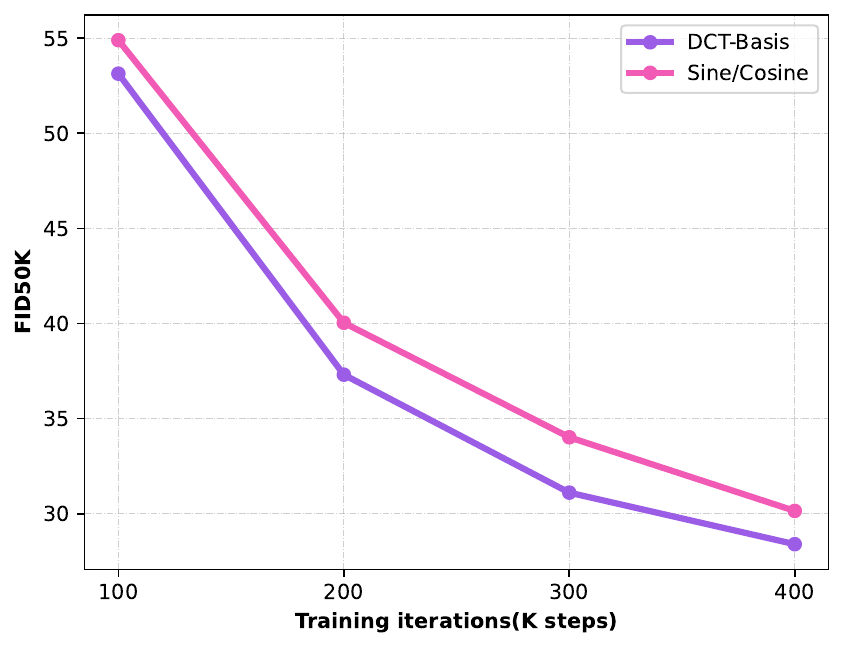}
    } 
    \subfloat[Interval Guidance\label{fig:interval_cfg}]{
        \includegraphics[width=0.32\linewidth]{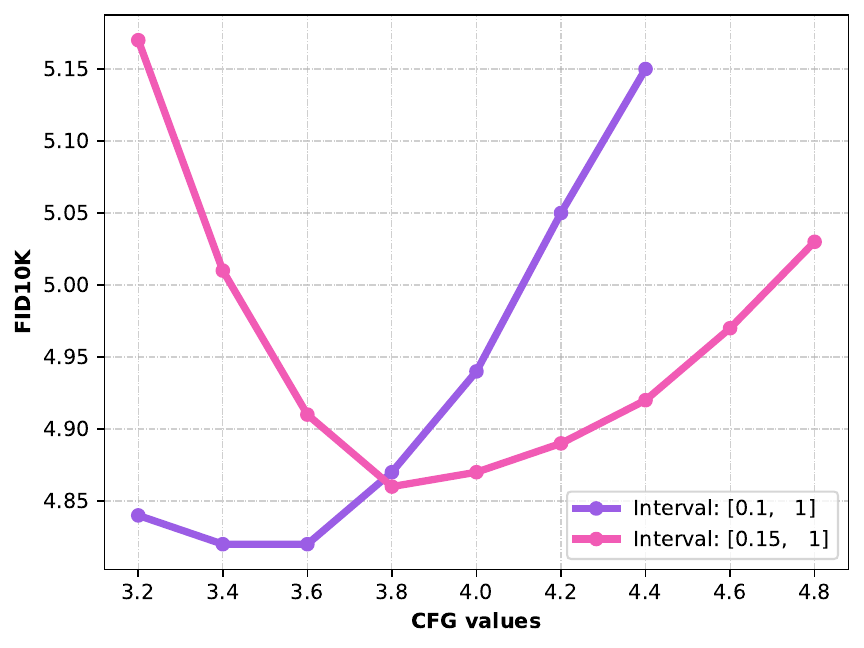}
    } 
    \subfloat[Sampling Solver\label{fig:adams}]{
        \includegraphics[width=0.32\linewidth]{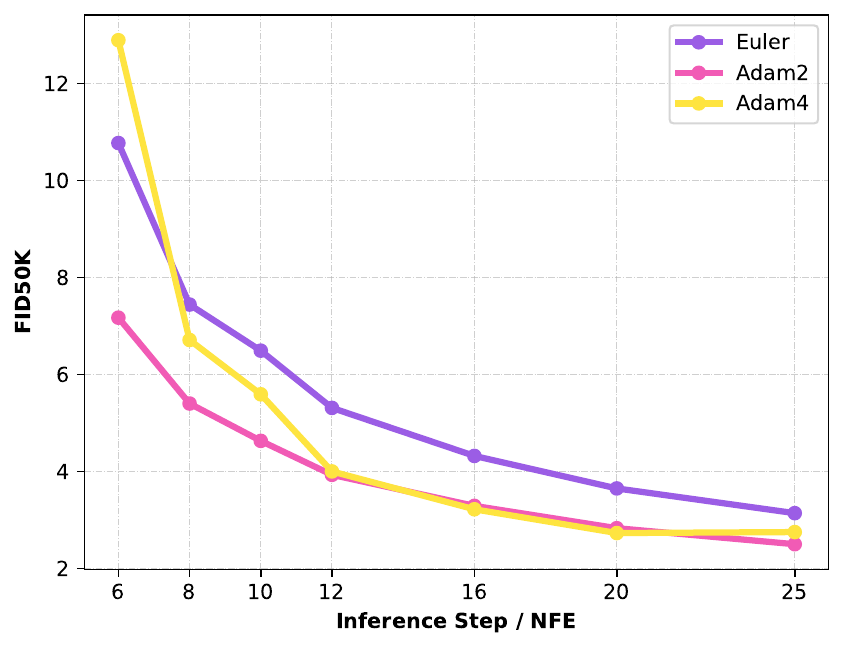}
    } 
    \caption{\textbf{Ablations studies of PixNerd.} {\small We conduct ablation studies on class-to-image generation benchmark ImageNet$256\times256$ with PixNerd-L/16.}}
    \vspace{-2em}
\end{figure}

\paragraph{Neural Field Normalization} As illustrated in \cref{fig:nerf_norm}, we evaluate different neural field normalization strategies. Our baseline compares three approaches: (1) normalizing only the first weight (FC1), (2) normalizing both weight (FC1/FC2), and (3) our default strategy that additionally normalizes the output features. Experimental results demonstrate that the default strategy achieves optimal performance and convergence speed.

\paragraph{Neural Field MLP channels} We conduct an empirical study of different MLP channel configurations (36, 64, and 72 channels) as shown in \cref{fig:nerf_channels}. Our experiments reveal that: (1) a minimal configuration of 36 channels leads to noticeable performance degradation compared to 64 channels; (2) while the 72-channel variant achieves marginally better results, it incurs significant computational overhead, including slower training speed and increased parameter count. Based on this trade-off analysis, we select 64 channels as our default configuration.

\paragraph{Neural Field MLP Depth.} We investigate the impact of neural field depth by evaluating PixNerd-L/16 with 1, 2, and 4 MLP layers, as shown in \cref{fig:nerf_depth}. Our experiments demonstrate consistent performance improvements with increasing network depth. However, considering the trade-off between computational efficiency (inference latency and training speed) and model performance, we establish the 2-layer configuration as our optimal default architecture.
\paragraph{Coordinate Encodings} We compared the DCT-Basis coordinate encoding with traditional sine/cosine encoding in \cref{fig:nerf_pe}. Our DCT-Basis encoding achieves much better results than sine/cosine encoding in terms of both convergence and final result.
\begin{equation}
    \text{DCT-PE}(i,j) = \{ \cos(k_1 i)\cos(k_2 j),  \}_{k_1, k_2 \in (0, K]} . \label{eq:dct_encodings}
\end{equation}

\subsection{Inference Scheduler design}
\paragraph{Interval Guidance} Classifier-free guidance~\cite{dit, sit, cfg} is a commonly used technique to improve the diffusion model performance. Interval guidance~\cite{interval_guidance} is an improved cfg technique, and has been validated by recent works~\cite{vavae, decoupled_dit}. We sweep different CFG values from 3.0 to 5.0 with a step of 0.2 to find the optimal CFG scheduler. As shown in \cref{fig:interval_cfg}, our PixNerd-XL/16 achieves best result FID10k result with 3.4 or 3.6 within the interval $[0.1, 1]$. So we take 3.5 as the default CFG value.
\paragraph{Sampling Solver} In \cref{fig:adams}, we armed PixNerd-XL/16 with an Euler solver and an Adams-like linear multistep solver with 2/4 orders. The Adams-2 order solver consistently achieves better results than the Euler solver with limited sampling steps. Due to the learning difficulty of pixel spaces, Adams-4-order solver performs unstable compared to the Euler and Adams2 solvers. However, with sufficient sampling steps, their performance gap becomes marginal.  

\subsection{Class-to-Image Generation}
\begin{table*}[t]
\centering
\small 
\begin{tabular}{c|c|c|ccccc}
\toprule
& \multicolumn{6}{c}{\textbf{ImageNet} 256$\times$256}\\
\toprule
& Params & Epochs
& FID$\downarrow$ & sFID$\downarrow$ & IS$\uparrow$ & Pre.$\uparrow$ & Rec.$\uparrow$ \\
\midrule
\multicolumn{7}{l}{\textit{Latent Generative Models}} \\
LDM-4~\cite{ldm} & 400M + 86M & 170 & 3.6 & - & 247.7 & {0.87} & 0.48 \\

DiT-XL~\cite{dit} & 675M + 86M & 1400 & 2.27 & 4.60  & 278.2 & {0.83} & 0.57 \\

SiT-XL~\cite{sit} & 675M + 86M & 1400 & 2.06 & 4.50 & 270.3 & 0.82 & 0.59 \\

FlowDCN~\cite{flowdcn} & 618M + 86M & 400 & 2.00 & 4.33 & 263.1 & 0.82 & 0.58 \\

REPA~\cite{repa} & 675M + 86M & 800 & 1.42 & 4.70 & 305.7 & 0.80 & 0.64 \\

DDT-XL~\cite{decoupled_dit} & { 675M + 86M} & {400} & {1.26} & 4.51 & {310.6} & {0.79} & {0.65} \\

{MAR-L~\cite{mar}} & {479M + 86M} & {800} & {1.78} & - & {296.0} & {0.81} & {0.60} \\
{CausalFusion~\cite{causalfusion}} & {676M + 86M} & {800} & {1.77} & - & {282.3} & {0.82} & {0.61} \\

\midrule
\multicolumn{7}{l}{\textit{Pixel Generative Models}} \\
ADM~\cite{adm} & 554M & 400  & 4.59 & 5.13 & 186.7 & 0.82 & 0.52  \\
RDM~\cite{rdm} & 553M + 553M& / & 1.99 & 3.99 & 260.4 & 0.81 & 0.58 \\
JetFormer~\cite{jetformer} & 2.8B & / & 6.64 & - & - & 0.69 & 0.56 \\
FractalMAR-H \cite{fractal} & 844M & 600 & 6.15 & - & 348.9 & 0.81 & 0.46 \\
PixelFlow-XL/4~\cite{pixelflow} & 677M & 320 & 1.98 & 5.83 & 282.1 &  0.81 & 0.60 \\

\textbf{PixNerd-L/16~(Euler-50)} & 458M & 160 & 2.64 & 5.25 & 297 & 0.78 & 0.60 \\
\textbf{PixNerd-XL/16~(Euler-50)} & 700M & 160 & 2.29 & 4.82 & 303 & 0.80 & 0.59 \\
\textbf{PixNerd-XL/16~(Adam2-50)} & 700M & 160 & 2.16 & 4.93 & 291 & 0.78 & 0.60 \\
\textbf{PixNerd-XL/16~(Euler-100)} & 700M & 160 & 2.15 & 4.55 & 297 & 0.79 & 0.59 \\
\bottomrule
\end{tabular}
\caption{
\textbf{System performance comparison} on ImageNet $256\times256$ class-conditioned generation. {\small Our PixNerd-XL/16 achieves comparable results with latent diffusion models under similar computation demands while achieving much better results than other pixel space generative models. We adopt the interval guidance with interval $[0.1, 1]$ and CFG of 3.5.}
}
\vspace{-2em}
\label{tab:imagenet256_sota}
\end{table*}
\begin{table}[t] 
\centering
\small 
\begin{tabular}{l|c|ccccc}
\toprule
& \multicolumn{5}{c}{\bf{ImageNet} $512\times512$} \\
\toprule
Model & Params & FID$\downarrow$  & sFID$\downarrow$  & IS$\uparrow$  & Pre.$\uparrow$ & Rec.$\uparrow$ \\
\midrule
\multicolumn{7}{l}{\textit{Latent Diffusion Models}} \\

{DiT-XL/2}~\cite{dit} & 675M + 86M & {3.04} & {5.02} & {240.82} & 0.84 & 0.54 \\
{SiT-XL/2} ~\cite{sit} & 675M + 86M & 2.62 &  4.18 & 252.21 & 0.84 & 0.57 \\
{REPA-XL/2} ~\cite{repa} & 675M + 86M  & 2.08 & 4.19 & 274.6 & 0.83 & 0.58 \\
{FlowDCN-XL/2} ~\cite{flowdcn} & 608M + 86M & {2.44} & {4.53} & {252.8} & 0.84 & 0.54 \\
EDM2 ~\cite{edm} & 1.5B + 86M& 1.81 & & & \\
{DDT-XL/2} ~\cite{decoupled_dit} & 675M + 86M & { 1.28} & { 4.22} & {305.1} & { 0.80} & { 0.63} \\
\midrule
\multicolumn{6}{l}{\textit{Pixel Diffusion Models}} \\
ADM-G~\cite{adm}& 559M & 7.72 & 6.57 & 172.71 & {0.87} & 0.42 \\
ADM-G, ADM-U & 559M & 3.85 & 5.86 & 221.72 & 0.84 & 0.53\\
RIN~\cite{rin} & 320M & 3.95 & - & 210 & - & - \\
SimpleDiffusion\cite{simple_diffusion} & 2B & 3.54 & - & 205 & - & -  \\
\textbf{PixNerd-XL/16 (Euler50)} & 700M & 3.41 & 6.43 & 246.45 & 0.80 & 0.58 \\
\textbf{PixNerd-XL/16 (Euler100)} & 700M & 2.84 & 5.95 & 245.62 & 0.80 & 0.59 \\

\bottomrule
\end{tabular}
\vspace{1em}
\caption{\textbf{Benchmarking class-conditional image generation on ImageNet 512$\times$512.} {\small Our PixNerd-XL/16($512\times512$) is fine-tuned from the same model trained on $256\times256$ resolution. We adopt the interval guidance with interval $[0.1, 1]$ and CFG of 3.5.}}
\label{tab:imagenet512_sota}
\vspace{-2em}
\end{table}

\paragraph{Training details} In class-to-image generation, to ensure a fair comparative analysis, we did not use gradient clipping and learning rate warm-up techniques. We adopt EMA with 0.9999 to stabilize the training. Our default training infrastructure consisted of $8\times$ A100 GPUs.

\paragraph{Visualizations.} We placed the selected visual examples from PixNerd-XL/16 trained on ImageNet$256\times256$ and ImageNet$512\times512$ at \cref{fig:teaser}. Our PixNerd-XL/16 can generate images with promising details. We generate these images with a CFG of 3.5 and the Euler-50 solver.

\paragraph{ImageNet $256\times256$ Benchmark.} We report the metrics of PixNerd-L/16 and PixNerd-XL/16 in \cref{tab:imagenet256_sota}. Under merely 160 training epochs, PixNerd-L/16 achieves 2.64 FID, significantly better than other pixel generative models like Jetformer~\cite{jetformer}, FractalMAR~\cite{fractal}. Further, PixNerd-XL/16 achieves 2.29 FID under 50 steps with the Euler solver. When used with Adams-2-order-solver(Adam2 for brevity), PixNerd-XL/16 achieves 2.16 FID, comparable to DiT. With enough sampling steps,  PixNerd-XL/16 boosts FID to 2.15 under 100 steps.We adopt the interval guidance with interval $[0.1, 1]$ and CFG of 3.5. 

\paragraph{ImageNet $512\times512$ Benchmark.} We provide the final metrics of PixNerd-XL/16 at \cref{tab:imagenet512_sota}. To validate the superiority of our PixNerd model, we take our PixNerd-XL/16 trained on ImageNet~$256\times256$ as the initialization, fine-tune our PixNerd-XL/16 on ImageNet~$512\times512$ for $abc$ steps. We adopt the aforementioned interval guidance~\cite{interval_guidance} and we achieved 2.84 FID, with CFG of 3.5 within the time interval $[0.3, 1.0]$. PixNerd-XL/16 achieves comparable performance to other diffusion models.

\subsection{Text-to-Image Generation}
\begin{table}
\centering
\small
\setlength{\tabcolsep}{1.2pt}
\begin{tabular}{ll|c|cccccc|c}
\toprule
& & & \multicolumn{7}{c}{\textbf{GenEval} Benchmark} \\
&  & \#Params  & Sin.Obj. & Two.Obj. & Counting & Colors & Pos & Color.Attr. & Overall \\
\midrule
\multicolumn{2}{l}{\hspace{-.5em} \textit{autoregressive model}}  \\
& Show-o~\cite{showo} & 1.3B& 0.95 & 0.52 & 0.49 & 0.82 & 0.11 & 0.28 & 0.53 \\
& MAR~\cite{fluid} & 1.1B+4.7B+86M & 0.96 & 0.77 & 0.61 & 0.78 & 0.34 & 0.53 & 0.67 \\
& SimpleAR~\cite{simplear} & 0.5B &  - & 0.82 & - & - & 0.26 & 0.38 & 0.59 \\
& SimpleAR~\cite{simplear} & 1.5B & - & 0.90 & - & - & 0.26 & 0.45 & 0.63 \\
\midrule
\multicolumn{2}{l}{\hspace{-.5em} \textit{latent diffusion model}} \\
& LDM\cite{ldm} & 1.4B  & 0.92 & 0.29 & 0.23 & 0.70 & 0.02 & 0.05 & 0.37 \\
& DALL-E 2 & 4.2B  & 0.94 & 0.66 & 0.49 & 0.77 & 0.10 & 0.19 & 0.52 \\
& DALL-E 3 & - &0.96 & 0.87 & 0.47 & 0.83 & 0.43 & 0.45 & 0.67 \\
& Imagen & \hspace{6pt}3B & - & - & - & - & - & - & - \\
& SD3~\cite{sd3} & \hspace{6pt}8B& 0.98 & 0.84 & 0.66 & 0.74 & 0.40 & 0.43 & 0.68 \\
& Transfusion~\cite{transfusion} & 7.3B & - & - & - & - & - & - & 0.63 \\
\midrule
\multicolumn{2}{l}{\hspace{-.5em} \textit{pixel diffusion model}} \\
& PixelFlow-XL/4~\cite{pixelflow} &  882M + 3B & - & - & - & - & - & - & 0.60 \\
& PixelFlow-XL/4$^\dagger$~\cite{pixelflow} & 882M + 3B & - & - & - & - & - & - & 0.64\\
\midrule
& PixelNerd-XXL/16 & 1.2B + 1.7B & 0.97 & 0.86 & 0.44 & 0.83 & 0.71 & 0.53 & 0.73 \\
\bottomrule
\end{tabular}
\vspace{1em}
\caption{\textbf{Comparsion with other text-to-image models on GenEval Benchmark.}{\small $^\dagger$ indicates prompt rewriting. Parameters consist of denoiser+text encoder+vae. Our PixNerd-XXL/16 achieves competitive performance compared with others under a much-limited data scale~(45M images).}}
\label{tab:geneval}
\vspace{-2em}
\end{table}
\begin{table}
\centering
\small
\setlength{\tabcolsep}{1.5pt}
\begin{tabular}{ll|c|ccccc|c}
\toprule
& & & \multicolumn{6}{c}{\textbf{DPG} Benchmark} \\

& Model & \#Params & Global & Entity & Attribute & Relation & Other & Average \\
\midrule
\multicolumn{2}{l}{\hspace{-.5em} \textit{latent diffusion model}} \\
& SD v2 & 0.86B + 0.7B+ 86M  & 77.67 & 78.13 & 74.91 & 80.72 & 80.66 & 68.09 \\
& PixArt-$\alpha$ & 0.61B + 4.7B + 86M  & 74.97 & 79.32 & 78.60 & 82.57 & 76.96 & 71.11 \\
& Playground v2 & 2.61B + 0.7B + 86M & 83.61 & 79.91 & 82.67 & 80.62 & 81.22 & 74.54 \\
& DALL-E 3 & -  & 90.97 & 89.61 & 88.39 & 90.58 & 89.83  & 83.50\\
& SD v1.5 & 0.86B + 0.7B + 86M & 74.63 & 74.23 & 75.39 & 73.49 & 67.81  & 63.18\\
& SDXL & 2.61B +1.4B + 86M & 83.27 & 82.43 & 80.91 & 86.76 & 80.41 & 74.65\\
\midrule
\multicolumn{2}{l}{\hspace{-.5em} \textit{pixel diffusion model}} \\
& PixelFlow-XL/4 & 882M + 3B  & - & - & - & - & - & 77.90\\ 
& PixelNerd-XXL/16 & 1.2B + 1.7B & 80.5 & 87.9 & 87.2 & 91.3 & 72.8  & 80.9\\
\bottomrule
\end{tabular}
\vspace{1em}
\caption{\textbf{Comparsion with other text-to-image models on DPG Benchmark.}{\small Parameters consist of denoiser+text encoder+vae. Our PixNerd-XXL/16 achieves competitive performance compared with others under a much-limited data scale~(45M images).}}
\label{tab:dpg}
\vspace{-2em}
\end{table}

\paragraph{Data preprocess details} For text-to-image generation, we trained our model on a mixed dataset containing approximately 45M images from open-sourced datasets, e.g., SAM~\cite{sam}, JourneyDB~\cite{jdb}, ImageNet-1K~\cite{imagenet}. We recaption all the images with Qwen2.5-VL-7B~\cite{qwen2vl} to yield English caption descriptions of various lengths. Note that our caption results only contain English descriptions. All the images are cropped into a square shape of $256\times256$ or $512\times512$, we do not adopt various aspect ratio training. We leave the native resolution~\cite{nit} or native aspect training~\cite{seedream2, seedream3, mogao} as future works.

\paragraph{Training details} We adopt Qwen3-1.7B\footnote{https://huggingface.co/Qwen/Qwen3-1.7B} as the text encoder. To improve the alignment of frozen text features \cite{fluid}, we jointly train several transformer layers on the frozen text features similar to Fluid~\cite{fluid}. To further enhance the generation quality, we adopt an SFT stage at resolution $512\times512$ with the dataset released by BLIP-3o~\cite{blip3o}. The total batch size is 1536 for $256\times256$ resolution pretraining and 512 for $512\times512$ resolution pretraining. We trained PixNerd on $256\times256$ resolution for 200K steps and trained on $512\times512$ resolution for 80K steps. The default training infrastructure consisted of $16\times$ A100 GPUs. We adopt the gradient clip to stabilize training. We adopted \textit{torch.compile} to optimize the computation graph to reduce the memory and computation overhead. We use the Adams-2nd solver with 25 steps as the default choice for sampling. 

\begin{figure}
    \includegraphics[width=1.0\linewidth]{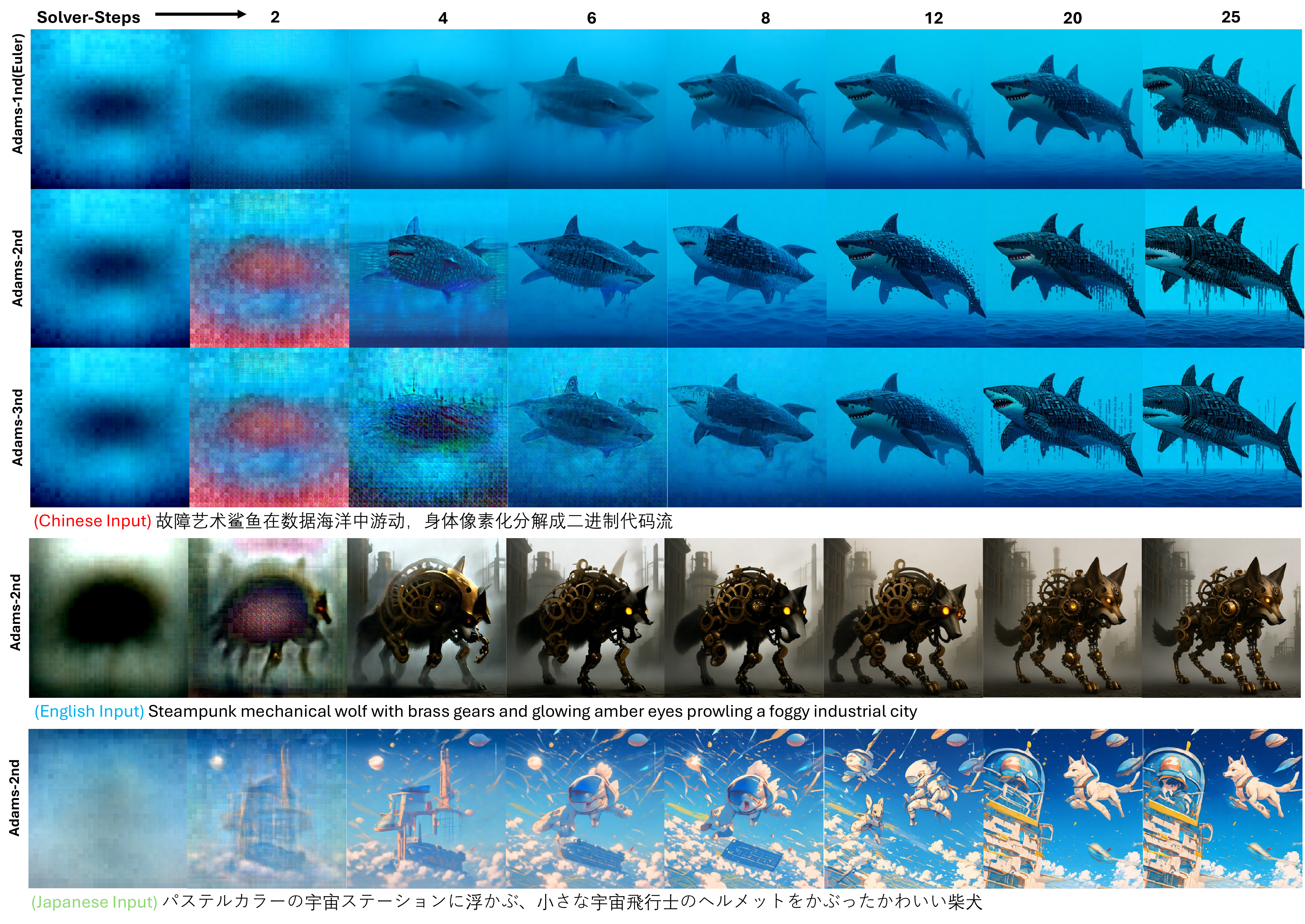}
    \caption{\textbf{The Text-to-Image $512\times512$ visualization with different solvers.} {\small We armed PixNerd with different ODE solvers, eg, Euler, Adams-2nd, Adams-3rd. Adams solver achieves better visual quality than the naive Euler solver. Also, thanks to the powerful text embedding in Qwen3 models, though we only trained PixNerd with English captions, PixNerd can generate samples of promising quality with other languages.}}
    \label{fig:t2i_vis_solver}
    \vspace{-1em}
\end{figure}

\paragraph{Visualizations.} We provided $512\times512$ visualizations with prompts provided by DouBao-1.5-Pro\footnote{https://www.volcengine.com/product/doubao} in \cref{fig:t2i_vis}. As illustrated in \cref{fig:t2i_vis}, our PixNerd-XXL/16 is capable of generating visually compelling images from complex text prompts. Overall, the atmosphere and color tones are largely accurate. Noted that we only trained PixNerd with English prompts. As shown in \cref{fig:t2i_vis_solver}, we can generate images even with other languages like Chinese and Japanese thanks to the powerful embedding space of Qwen3 models~\cite{qwen2vl}. Nevertheless, occasional blurry or unnatural artifacts appear in certain scenarios (e.g., steampunk lab image). We posit that appropriate post-training processing could mitigate such artifacts~\cite{hypersd}, and we intend to explore pixel-space post-training as future work.

\paragraph{Sampling solvers.} We provide denoising trajectories in $x_1$ space (clean data) of different solvers in \cref{fig:t2i_vis_solver}, including Euler, Adams-2nd, and Adams-3rd solvers. Adams-2nd solver achieves stable and fast sampling results. 

\paragraph{Training-free arbitrary resolution generation.} As shown in \cref{fig:t2i_vis_multires}, without any resolution adaptation fine-tuning, we can achieve arbitrary resolution generation through the coordinate interpolation while keeping the amount of tokens as constant as the pretraining resolution. Specifically, we sampled pretraining resolution from the given noisy image, then fed this sampled version into the transformer. This keeps the token amount in our PixNerd as consistent as in the pretraining stage.  To match the velocity field with the desired resolution of the given noisy image, we then interpolate the coordinates for neural field decoding accordingly.

\paragraph{GenEval Benchmark} We provided quantity comparison on Geneval~\cite{geneval} benchmark in \cref{tab:geneval}. Our PixNerd-XXL/16 achieves comparable results under enormous patch sizes and limited data scales. As shown in \cref{tab:geneval}, our PixNerd-XXL/16 achieves 0.73 overall score, beating pixelflow~\cite{pixelflow} with a significant margin.

\paragraph{DPG Benchmark} We provided quantity comparsion on DPG~\cite{dpg} benchmark in \cref{tab:dpg}. Our PixNerd-XXL/16 achieves competitive results compared to its latent counterparts. As shown in \cref{tab:dpg}, our PixNerd-XXL/16 achieves 0.82 overall score, beating other pixel generation models with a significant margin.

\section{Discussion}
Several related works~\cite{infd, ddmi, inrflow} also combine diffusion with neural fields. In practice, however, they differ fundamentally from PixNerd. For example, INFD~\cite{infd} and DDMI~\cite{ddmi} leverage neural fields to enhance VAEs rather than diffusion models, and their generative capacity still stems from a latent diffusion model. DenoisedWeights~\cite{denoised_weights} trains independent neural weights for each image before training a generative model on these pre-collected weights. This remains a two-stage framework and poses non-trivial challenges for large-scale training. INRFlow~\cite{inrflow} and PatchDiffusion\cite{patchdiffusion} utilize coordinate encodings to enhance diffusion model performance.  Beyond diffusion-based generative models, GAN-based methods~\cite{asis, agci, coco_gan, stylegan3} also utilize neural fields or coordinate encodings. PixNerd is a simple yet elegant single-stage pixel-space generative model that does not rely on a VAE. Current latent generative models inevitably cascade errors due to their two-stage configurations. Further, a high-quality VAE usually demands numerous losses supervisions, e.g., adversarial loss, LPIPS loss. In particular, adversarial loss is unstable in training and tends to introduce artifacts. Pixel generative model has more potential in the future, and PixNerd is a simple yet elegant solution for a pixel-space generative model.

\section{Conclusion}
In this paper, we return to pixel space diffusion with neural field. We present a single-scale, single-stage, efficient, end-to-end solution, the pixel neural field diffusion~(PixelNerd). We achieved 2.15 FID on ImageNet $256\times256$ and 2.84 FID on ImageNet $512\times512$ without any complex cascade pipeline. Our PixNerd-XXL/16 achieved a competitive 0.73 overall score on the GenEval benchmark and 80.9 average score on DPG benchmark. However, current PixNerd shows unclear details in some cases, as \cref{fig:t2i_vis} and still has gaps with its latent counterparts.

{
\small
\bibliographystyle{unsrt}
\bibliography{neurips_2025}
}

\newpage

\end{document}